\documentclass[conference]{IEEEtran}
\usepackage{cite}
\ifCLASSINFOpdf
   \usepackage[pdftex]{graphicx}
\else
   \usepackage[dvips]{graphicx}
\fi

\usepackage{amsmath}

\usepackage{array}

\usepackage{url}
\usepackage{hyperref}
\usepackage{stfloats}

\usepackage{fancyhdr}
\usepackage{stfloats}
\begin{document}

\title{Hyperspectral Image Dataset\\for Benchmarking on Salient Object Detection}

\author{\IEEEauthorblockN{Nevrez Imamoglu\IEEEauthorrefmark{1},
Yu Oishi\IEEEauthorrefmark{1},
Xiaoqiang Zhang\IEEEauthorrefmark{2}, 
Guanqun Ding\IEEEauthorrefmark{2},
Yuming Fang\IEEEauthorrefmark{2},\\
Toru Kouyama\IEEEauthorrefmark{1} and Ryosuke Nakamura\IEEEauthorrefmark{1}}
\IEEEauthorblockA{\IEEEauthorrefmark{1}Artificial Intelligence Research Center, National Institute of Advanced Industrial Science and Technology,
Tokyo, Japan\\ Email (corresponding author): nevrez.imamoglu@aist.go.jp or nevrez@ieee.org}
\IEEEauthorblockA{\IEEEauthorrefmark{2}School of Information Technology, Jiangxi University of Finance and Economics, Nanchang, China}}

\maketitle

\thispagestyle{fancy}
\begin{abstract}
Many works have been done on salient object detection using supervised or unsupervised approaches on colour images. Recently, a few studies demonstrated that efficient salient object detection can also be implemented by using spectral features in visible spectrum of hyperspectral images from natural scenes. However, these models on hyperspectral salient object detection were tested with a very few number of data selected from various online public dataset, which are not specifically created for object detection purposes. Therefore, here, we aim to contribute to the field by releasing a hyperspectral salient object detection dataset with a collection of 60 hyperspectral images with their respective ground-truth binary images and representative rendered colour images (sRGB). We took several aspects in consideration during the data collection such as variation in object size, number of objects, foreground-background contrast, object position on the image, and etc. Then, we prepared ground truth binary images for each hyperspectral data, where salient objects are labelled on the images. Finally, we did performance evaluation using Area Under Curve (AUC) metric on some existing hyperspectral saliency detection models in literature. 
\end{abstract}


% \fi
\IEEEpeerreviewmaketitle

\section{Introduction}

Visible spectrum may contain more information than that of the colour images captured by most end-user cameras with three spectral measurements (Red-Green-Blue) for scene analysis or computer vision applications \cite{1}. Hyperspectral images obtained from spectral cameras provides higher spectral resolution and have spectral information on several narrow spectral bands at each pixel \cite{1,2,3,4}. Many applications in various fields (e.g. remote sensing, computer vision) have taken advantage of the spatial and spectral information of hyperspectral cameras \cite{1}; for example, in applications such as remote sensing \cite{5,6,7}, scene/object analysis or object detection \cite{3,4,5,6,7,8,9}, spectral estimation \cite{9,10,11,12}, etc.

One of the possible applications of hyperspectral imagery can be salient object detection in natural scenes based on the visual attention mechanism, in which algorithms aim to explore objects or regions more attentive than the surrounding areas on the scene or images \cite{3,13,14}. The first computational model of saliency detection was proposed by Itti et al. \cite{13}, which takes advantage of center-surround differences on intensity, colour and orientation features in multi-scale. Following the work \cite{13}, many works have been done on salient object detection on colour or gray images for various supervised or unsupervised applications as in \cite{14,15,16,17}. 

Recently, a few studies \cite{3,4,7,8} demonstrated that efficient salient object detection can also be implemented by using spectral features in visible spectrum of hyperspectral images from natural scenes. Most of these models, combine Itti et al. \cite{13} based center-surround differences with spectral features such as spectral angle similarity or similar saliency extraction features \cite{3,4,7,8}. Regarding the evaluation data, in \cite{7}, the application is remote sensing (aerial/satellite) data, which is not the target data of this work and salient objects are labelled with bounding boxes from images. In \cite{3}, only 13 hyperspectral images (31 spectral channels with 10nm intervals in 400nm - 700 nm visible range) were used, and ground truth for salient objects were done by labelling with bounding-boxes.  Yan et al. \cite{4} use similar sources with \cite{3} to collect data and evaluate their spectral gradient based saliency model, which are from publicly available online sources, and they improve the evaluation dataset by increasing the number to 17 images and labelling salient object with object boundaries rather than bounding box. 

In summary, these models \cite{3,4,8} on hyperspectral salient object detection were tested with a very few number of data selected from various online public dataset, which are not specifically created for object detection purposes. Therefore, this work aims to create a collection of larger hyperspectral image dataset from outdoor scenes that can be used for salient object detection task on hyperspectral data cubes. We aim to contribute to the field by releasing a salient object detection dataset with a collection of 60 hyperspectral images with their respective ground-truth binary images and representative rendered colour images(sRGB). We took several aspects in consideration during the data collection such as variation in object size, number of objects, foreground-background contrast, object position on the image, and etc. Then, we prepared ground truth binary images for each hyperspectral data, where salient objects are labelled with object boundaries on the images. Finally, we did performance evaluation using Area Under Curve (AUC) metric on existing hyperspectral saliency detection models from literature.

In the following section, we will explain the details of data collection process and data specifications. Then, we will demonstrate some results on the performance of salient object detection algorithms that are applicable on our dataset.

\section{Hyperspectral image dataset:\\for salient object detection benchmarking}

In this section, we will explain the hyperspectral image dataset for salient object detection. The dataset will be available on updated forked repository \href{https://github.com/nevrez/HS-SOD}{https://github.com/nevrez/HS-SOD} and \href{https://huggingface.co/datasets/gsvrg/HS-SOD}{https://huggingface.co/datasets/gsvrg/HS-SOD}
(Old repository \href{https://github.com/gistairc/HS-SOD}{https://github.com/gistairc/HS-SOD}). For data collection, NH-AIK model hyperspectral camera is used, which is based on NH-series (NH-5) \cite{18} (see Fig.1) and produced by Eba-Japan Co. Ltd \cite{18}. In Table \ref{table:table1}, specifications of the camera are given. 

\begin{figure}[h]
\centering
\includegraphics[width=0.40\textwidth]{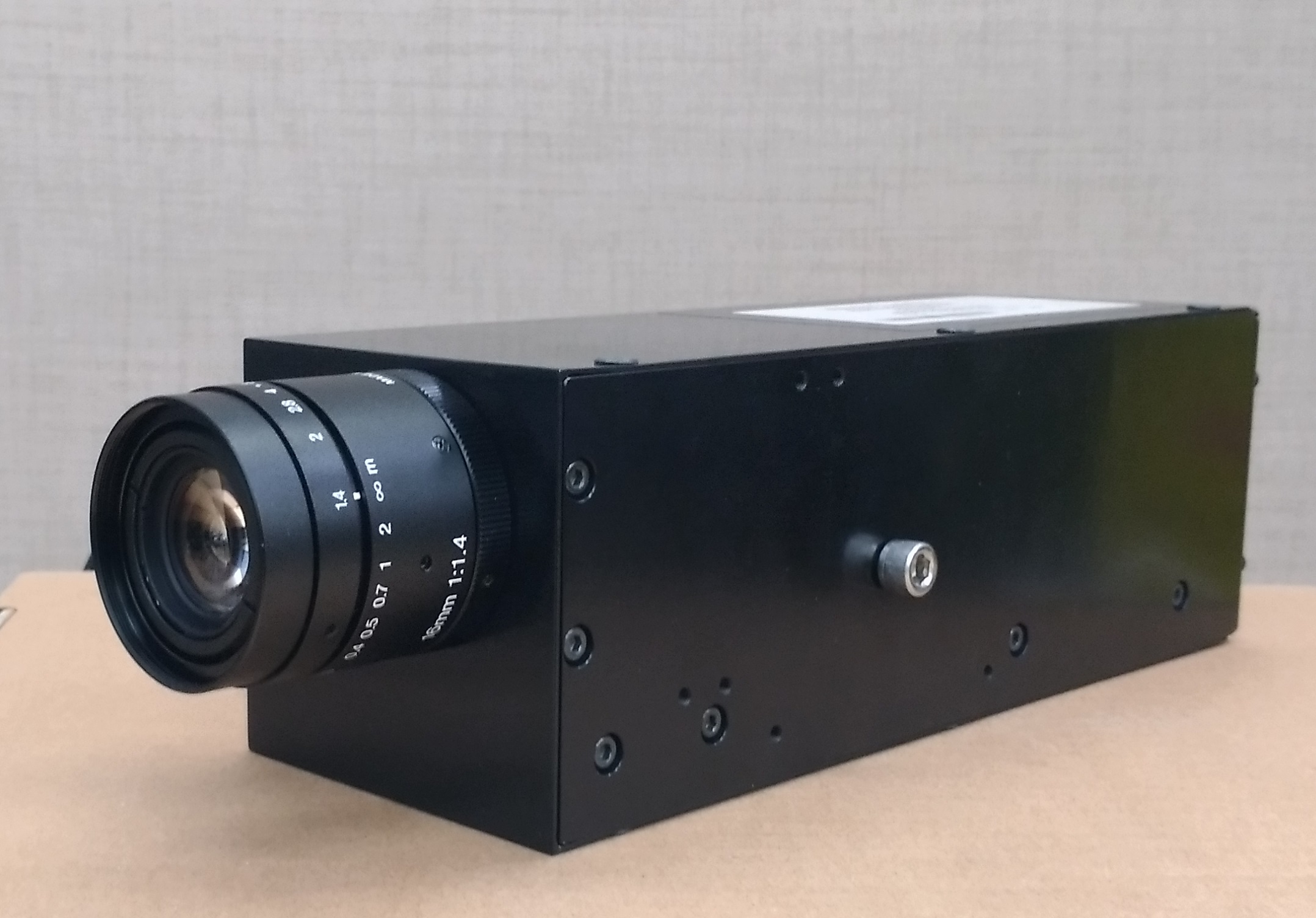}
\caption{NH-AIK Hyperspectral camera used for data collection}
\label{figcamera}
\end{figure}

\begin{table}[ht]
\caption{NH-AIK Hyperspectral Camera Properties}
\label{table:table1}
\centering
\begin{tabular}{|c||c|}
\hline
Image resolution & 1024 x 768 pixels\\
\hline
Measuring Wavelength & 350 - 1100 nm\\
\hline
Wavelength resolution & 5 nm\\
\hline
Number of spectral bands & 151 channels\\
\hline
File format & 10 bit, uint16,  Band Interleaved by Line\\
\hline
\end{tabular}
\end{table}

The data is collected at the public parks of Tokyo Waterfront City in Odaiba, Tokyo, Japan (see the green areas in Fig.2 \cite{19}) with the permission of Tokyo Port Terminal Corporation \cite{20}. We collected data in several days between August - September 2017 when the weather is sunny or partially cloudy. At each data collection day, a tripod was used to fix camera to minimize motion distortion on the images. We tried to keep the exposure time and gain for camera settings fixed as much as possible depending on the daylight conditions while keeping saturation of pixels values or image visibility in mind. As a reference to the dataset users, we are providing camera settings such as exposure time and gain values for each image in a text file with the corresponding data. We also did not apply normalization on captured bands. It may improve the quality of the hyperspectral images with higher colour contrast between foreground and background regions; however, it may also decrease the difficulty of dataset for benchmarking on salient object detection task. 
 
\begin{figure}[h]
\centering
\includegraphics[width=0.40\textwidth]{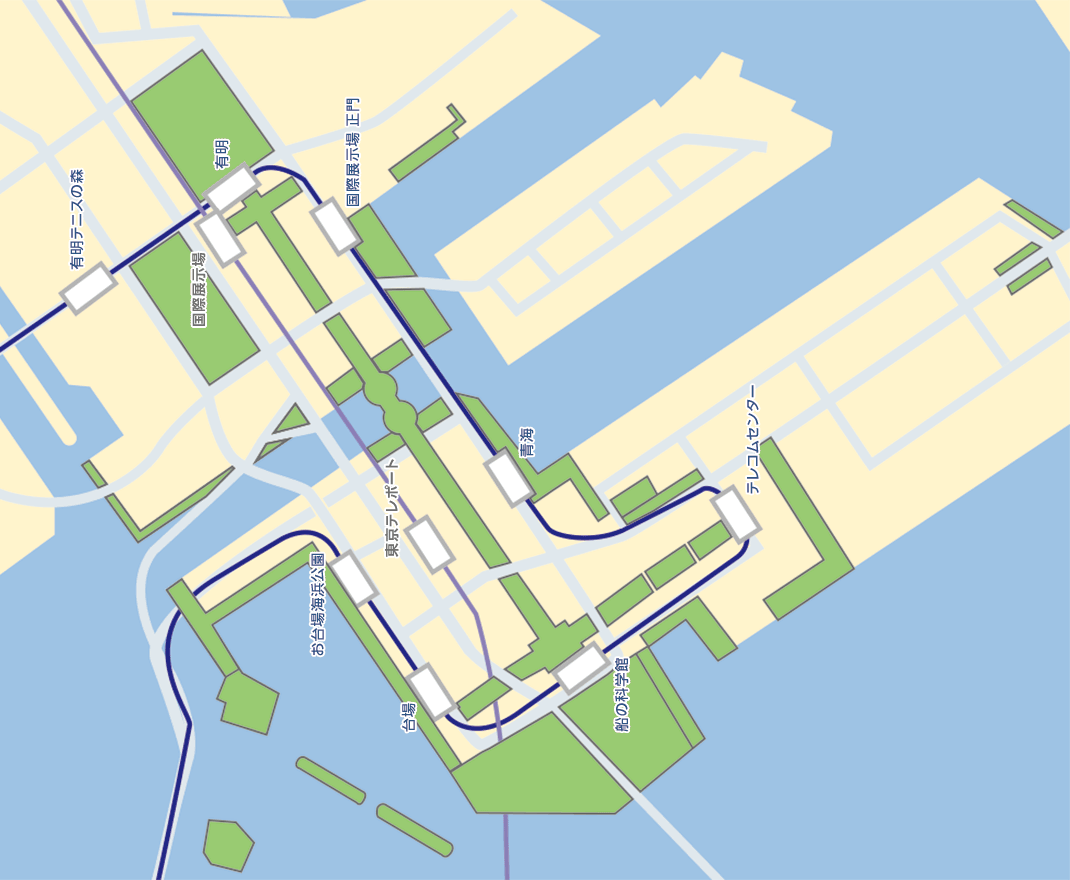}
\caption{Parks (green areas) at Tokyo Waterfront City (Odaiba, Tokyo) visited for data collection}
\label{figdatacollectionsite}
\end{figure}

After obtaining various hyperspectral images, we have selected 60 images from approximately fifty different scenes with the conditions: i) we removed distorted images due to motion in the scene (depending on the exposure time, one image may take a few seconds for camera), ii) we considered several aspects such as variations in salient object size, spatial positions of objects on images, number of salient objects, foreground-background contrast , iii) a few images has the same scene but the object positions, object distance, or number of objects varied. 

\begin{figure}[h]
\centering
\includegraphics[width=0.45\textwidth]{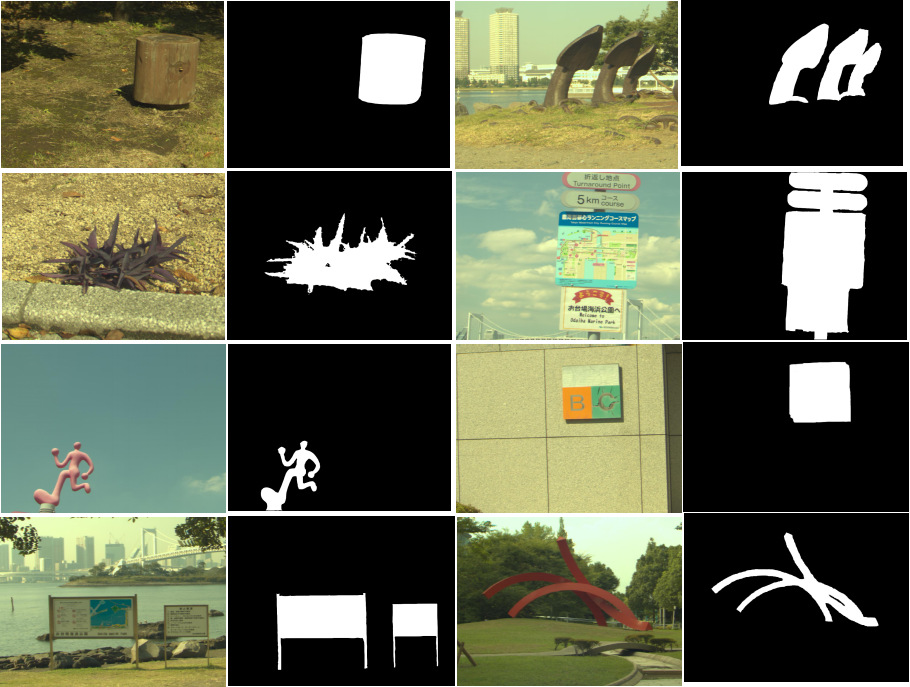}
\caption{Sample images of scenes from hyperspectral dataset rendered in sRGB and respective ground truth binary images for salient objects}
\label{figdatacollectionsite}
\end{figure}

For the convenience of salient object detection task, we cropped spectral bands around the visible spectrum and we saved hyper-cubes for each scene in ".mat" file format after sensor dark-noise correction. As defined in \cite{21}, visible spectrum has a well accepted range of 380 - 780 nm though the range between 400 - 700nm as in \cite{3,4} may also be used. To keep the range wide and flexibility to the people who want to use the dataset, we selected the defined range of 380 - 780 nm in \cite{21} for our dataset though visual stimulus might be weaker at the boundary of these ranges for human visual system \cite{21}. Then, we rendered in sRGB colour images from hyperspectral images to create ground-truth salient object binary images by labelling the boundaries of salient objects. In Fig.3, some example images are given from our hyperspectral dataset rendered in sRGB colour images with their respective ground truth binary images for salient objects.

\section{Experiments on dataset}

In this section, we test the dataset with spectral saliency models presented in \cite{3} and \cite{4}. For quantitative evaluation of the salient object detection performances, Area Under Curve (AUC) metric is selected; AUC implementation of Borji et al. \cite{22} (AUC-Borji) is used in our experiments. AUC is a commonly used metric for the comparison of salient object detection methods. The results are given in Table II.

We started out experiments by utilizing saliency computation from \cite{3}, in which the code of \cite{3} demonstrates various usage of spectral data for saliency detection. First of all, as a baseline model, saliency maps from Itti et al. \cite{13} were also computed for comparison in the code of \cite{3}. Then, the work in \cite{3} check spectral distances between each spatial region for saliency computation by using spectral Euclidean distance (SED) and spectral Angle distances (SAD). Also, in \cite{3}, colour opponency method in \cite{13} is replaced by spectral information rather than Red-Green and Blue-Yellow differences. To do compute saliency from spectral group (GS), spectral bands are divided into four groups (G1,G2,G3,G4), and then Euclidean distance between these vectors (G1-G3 and G2-G4) as colour opponency are calculated rather than single value colour component \cite{3}. In \cite{3}, orientation based salient features (OCM) are also adopted from \cite{13}. As in \cite{3}, the combinations SED-OCM-GS and SED-OCM-SAD were also tested on our dataset. From the various spectral saliency approaches in \cite{3},  SED-OCM-SAD yielded best AUC performance by giving 0.8008.

As a more recent work, we also tested saliency from spectral gradient contrast (SGC) proposed by \cite{4}. It should be noted that we implemented the model in \cite{4} since the codes were not available yet. In \cite{4}, local region contrast is computed from the super-pixels, where super pixels are obtained by considering both spatial and spectral gradients. SGC \cite{4} gives the best AUC performance on our dataset among the tested models by having 0.8205 AUC performance.

\begin{table}[ht]
\caption{Evaluation of spectral salient object detection methods on our dataset }
\label{table:table2}
\centering
\begin{tabular}{|c||c|}
\hline
State-of-the-Art Saliency Methods & AUC-Borji \cite{22} Performance\\
\hline
Itti et al \cite{13} & 0.7694\\
\hline
SED \cite{3} & 0.6415\\
\hline
SAD \cite{3} & 0.7521\\
\hline
GS \cite{3} & 0.7597\\
\hline
SED-OCM-GS \cite{3} & 0.7863\\
\hline
SED-OCM-SAD \cite{3} & 0.8008\\
\hline
SGC \cite{4} & 0.8205\\
\hline
\end{tabular}
\end{table}

\section{Conclusion}

 In this work, we presented a collection of larger hyperspectral image dataset (60 images with respective salient object ground-truths) that can be used for salient object detection task. Then, we tested our hyperspectral data with some spectral saliency models from \cite{3} and \cite{4}. Regarding, salient object detection task, SGC \cite{4} seems to be more robust compared to models in \cite{3}, probably, due to two main reasons; i) using region contrast may be less noisy than pixel-wise saliency, ii) spectral gradient may have higher invariance to illumination changes as stated in \cite{4}. However, despite being better than base line model \cite{13},  these initial spectral saliency results shows that there are still many things that can be proposed to improve spectral salient object detection performances since current AUC performances still does not seem to be at the level of state-of-the-art colour image based salient object detection methods. We hope that this dataset will help to improve research in this area.


\section*{Acknowledgement}
This paper based on the results obtained from a project commissioned by the New Energy and Industrial Technology Development Organization (NEDO).

\end{document}